\documentclass[runningheads]{llncs}

\usepackage{graphicx}
\usepackage{color}
\usepackage{epsfig}
\usepackage{amsmath}
\usepackage{amssymb}
\usepackage{multirow}
\usepackage[abs]{overpic}
\usepackage[breaklinks,colorlinks]{hyperref} 
\usepackage{url}            
\usepackage{booktabs}       
\usepackage{amsfonts}       
\usepackage{nicefrac}       
\usepackage{microtype}      
\usepackage{xcolor}         
\usepackage{cite}
\usepackage{caption}
\usepackage{makecell}
\usepackage{bbding} 
\usepackage{lipsum}
\usepackage{bm}
\usepackage[american]{babel}
\usepackage{subfloat}
\usepackage{float}
\usepackage{subfig}
\usepackage{tabularx}
\usepackage{adjustbox}
\usepackage{wrapfig}
\RequirePackage{xspace}
\usepackage[misc]{ifsym} 

\newcommand{\tabincell}[2]{\begin{tabular}{@{}#1@{}}#2\end{tabular}}

\makeatletter
\DeclareRobustCommand\onedot{\futurelet\@let@token\@onedot}
\def\@onedot{\ifx\@let@token.\else.\null\fi\xspace}

\def\ie{\emph{i.e}\onedot}

\def\etal{\emph{et al}\onedot}
\makeatother

\begin{document}
\title{RBSR: Efficient and Flexible Recurrent Network for Burst Super-Resolution}

\author{Renlong Wu\inst{} \and
Zhilu Zhang\inst{} \and
Shuohao Zhang\inst{} \and \\
Hongzhi Zhang\inst{} $^{(\textrm{\Letter})}$
\and
Wangmeng Zuo\inst{}
}
\authorrunning{Wu. et al.}
\institute{Harbin Institute of Technology, Harbin, China \\
\email{hirenlongwu@gmail.com,cszlzhang@outlook.com,\\yhyzshrby@163.com,zhanghz0451@gmail.com,wmzuo@hit.edu.cn}}
\titlerunning{RBSR}

\maketitle              
\begin{abstract}
Burst super-resolution (BurstSR) aims at reconstructing a high-resolution (HR) image from a sequence of low-resolution (LR) and noisy images, which is conducive to enhancing the imaging effects of smartphones with limited sensors. The main challenge of BurstSR is to effectively combine the complementary information from input frames, while existing methods still struggle with it. In this paper, we suggest fusing cues frame-by-frame with an efficient and flexible recurrent network. In particular, we emphasize the role of the base-frame and utilize it as a key prompt to guide the knowledge acquisition from other frames in every recurrence. Moreover, we introduce an implicit weighting loss to improve the model's flexibility in facing input frames with variable numbers. Extensive experiments on both synthetic and real-world datasets demonstrate that our method achieves better results than state-of-the-art ones. Codes and pre-trained models are available at \url{https://github.com/ZcsrenlongZ/RBSR}. 

\end{abstract}
\section{Introduction}
The rising prevalence of smartphones has led to an increasing demand for capturing high-quality images.
However, inherent disadvantages of the smartphone camera are inevitable in order to integrate it into the thin-profile device, including tiny sensor size, fixed aperture, and restricted zoom~\cite{delbracio2021mobile}.
Such physical demerits not only result in limited image spatial resolution, but also easily include  noise~\cite{Tsai1984MultiframeIR,dudhane2022burst}, especially in the low-light environment.
With the development of deep learning~\cite{he2016deep,liang2021swinir,vaswani2017attention,ronneberger2015u}, many single-image denoising~\cite{zhang2017beyond,zamir2022restormer,li2023spatially,wang2022uformer,chen2021pre} and super-resolution~\cite{dong2015image,liu2020deep,zhang2021learning,wang2021real} methods have attempted to address the issue, achieving great progress. 
Nevertheless, they are severely ill-posed and difficult to restore high-quality images with realistic and rich details when dealing with badly degraded images.
Instead, burst super-resolution (BurstSR)~\cite{bhat2021deep, bhat2021ntire, bhat2022ntire} can relax the ill-posedness and give a chance to bridge the imaging gap with DSLRs, by reconstructing a high-resolution (HR) sRGB image from continuously captured low-resolution (LR) and noisy RAW frames.

Two critical processes for BurstSR are inter-frame alignment and feature fusion~\cite{bhat2021deep,bhat2021mfir,dudhane2022burst,luo2022bsrt,luo2021ebsr,mehta2023gated,dudhane2023burstormer}.
The former has been more fully explored in various ways, involving the optical flow~\cite{sun2018pwc} alignment~\cite{chan2021basicvsr,bhat2021deep,bhat2021mfir}, deformable convolution~\cite{dai2017deformable} alignment~\cite{wang2019edvr,luo2022bsrt,mehta2023gated}, and cross-attention alignment~\cite{liang2022recurrent}.
But for the latter, existing approaches still have certain limitations in merging complementary information from multiple frames. 
Among those, the weighted-based fusion \cite{bhat2021deep,bhat2021mfir} combines aligned features by predicting element-wise weights, which only pays attention to the communication between base-frame and non-base frames, while ignoring information exchange among non-base frames.
The pseudo-burst fusion \cite{dudhane2022burst}  concatenates burst features channel by channel, being more sufficient yet only processing input frames with a fixed number.
The attention-based fusion \cite{luo2021ebsr,mehta2023gated,dudhane2023burstormer}  exploits inter-frame cross-attention mechanism to enhance feature interaction on the channel or spatial dimension, but it is computationally extensive.
Thus, it is still worth exploring a multi-frame fusion method that can be efficient and adaptive to variable numbers of frames in BurstSR.

\begin{figure}[t!]
    \centering
    \includegraphics[width=0.7\linewidth]{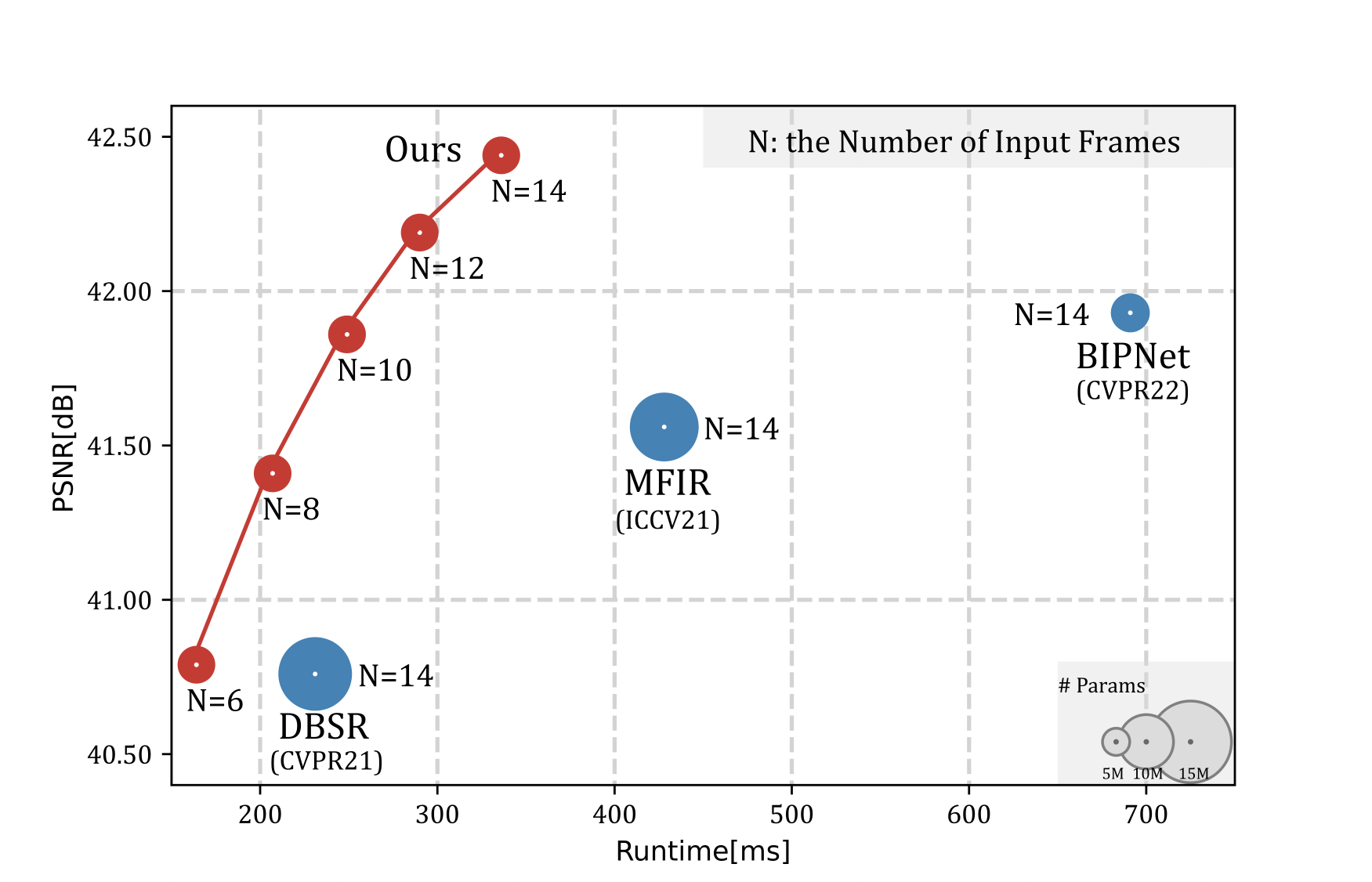}
    \vspace{-3mm}
    \caption{Performance and runtime comparison on SyntheticBurst dataset~\cite{bhat2021deep}.}
    \label{fig:PSNR_TIME}
    \vspace{-6mm}
\end{figure}
In this work, we suggest fusing cues frame-by-frame, which can specifically merge the beneficial information of each frame and has no limit on the number of frames.
A recurrent manner is a natural and suitable choice, and we propose an efficient and flexible recurrent network for burst super-resolution, dubbed RBSR.
RBSR processes the base-frame first and aggregates other aligned temporal features sequentially to reconstruct the HR result. 
In particular, we find that the base-frame plays an  important role in such a recurrent approach. 
Thus, we emphatically utilize the base-frame as a key prompt to guide the acquisition of knowledge from other frames in every recurrence.
Moreover, although the recurrent network itself has the ability to deal with variable lengths of inputs, simply constraining the output during the last recurrence will lead to a flexibility decrease in handling fewer frames.
Thus, we further propose an implicit weighting loss to enhance the model's ability in facing fewer frames while maintaining the performance in processing the longest frames. 

Experiments are conducted on both synthetic and real-world datasets~\cite{bhat2021deep}.
Benefiting from the compact and efficient method design, the results show that our RBSR not only achieves better fidelity as well as perceptual performance than state-of-the-art methods, but also has favorable efficiency.
In comparison with BIPNet~\cite{dudhane2022burst}, our RBSR obtains 0.51 dB PSNR gain while only taking less than half of its inference time, as shown in Fig.~\ref{fig:PSNR_TIME}.

The main contributions can be summarized as follows:

1. We focus on the efficient and flexible fusion manner in BurstSR, and propose a recurrent network named RBSR, where the base-frame is emphatically utilized to guide the knowledge acquisition from other frames in every recurrence by the suggested KFGR module. 

2. An implicit weighting loss is introduced to further enhance the model's ability in facing input frames with variable numbers.

3. Experiments on both synthetic and real-world datasets demonstrate that our method not only outperforms state-of-the-art methods quantitatively and qualitatively, but also has favorable inference efficiency.

\section{Related Work}
\noindent{\textbf{Single Image Super-resolution.}}
With the development of deep learning~\cite{he2016deep,liang2021swinir,vaswani2017attention,ronneberger2015u}, single-image super-resolution (SISR) has achieved great success in terms of both performance\cite{dong2015image,lim2017enhanced,zhang2020texture,zhang2018learning} and efficiency\cite{wang2021exploring,liu2020deep,kong2021classsr,xie2021learning,luo2022adjustable}.
SISR-oriented network designs~\cite{lim2017enhanced,kim2016accurate,lai2017deep,wei2020component} and optimization objectives~\cite{johnson2016perceptual,lugmayr2020srflow,kim2016accurate} are widely explored.
However, these methods still struggle with heavily degraded images and sometimes generate artifacts due to the severe ill-posedness of SISR.

\noindent{\textbf{Multi-Frame Super-resolution.}
Compared to SISR methods, multi-frame super-resolution (MFSR) approaches\cite{Tsai1984MultiframeIR,deudon2020highres,bhat2021deep,bhat2021mfir,dudhane2022burst,wronski2019handheld,mehta2023gated,dudhane2023burstormer,luo2021ebsr} aggregate multiple aliased images for better reconstruction.
Tsai\cite{Tsai1984MultiframeIR} first proposes a frequency domain method with the assumption that the input image translations are known. 
HighResNet\cite{deudon2020highres} focuses on satellite imagery, aligning each frame to a reference frame implicitly and performing recursive fusion. 
DBSR \cite{bhat2021deep} introduces a weighted-based fusion mechanism, where element-wise weights between the base-frame and other frames are predicted. 
MFIR\cite{bhat2021mfir} further extends this fusion mechanism by learning the image formation model in deep feature space. 
However, only paying attention to communication between the base-frame and non-base frames limits complementary information exchange among non-base ones.
Recently, BIPNet \cite{dudhane2022burst} proposes a pseudo-burst fusion strategy by fusing temporal features channel-by-channel, being more sufficient but requiring fixing the input frame number.
A few works\cite{luo2021ebsr,mehta2023gated,dudhane2023burstormer} exploits inter-frame attention-based mechanism to enhance feature interaction via cross-attention on the channel or spatial dimension, yet are computationally extensive.
In this work, we suggest fusing cues frame-by-frame, which enables sufficient integration of advantageous information from each frame and is not constrained by the number of frames.
\begin{figure*}[h]
    \centering
    \includegraphics[width=0.8\linewidth]{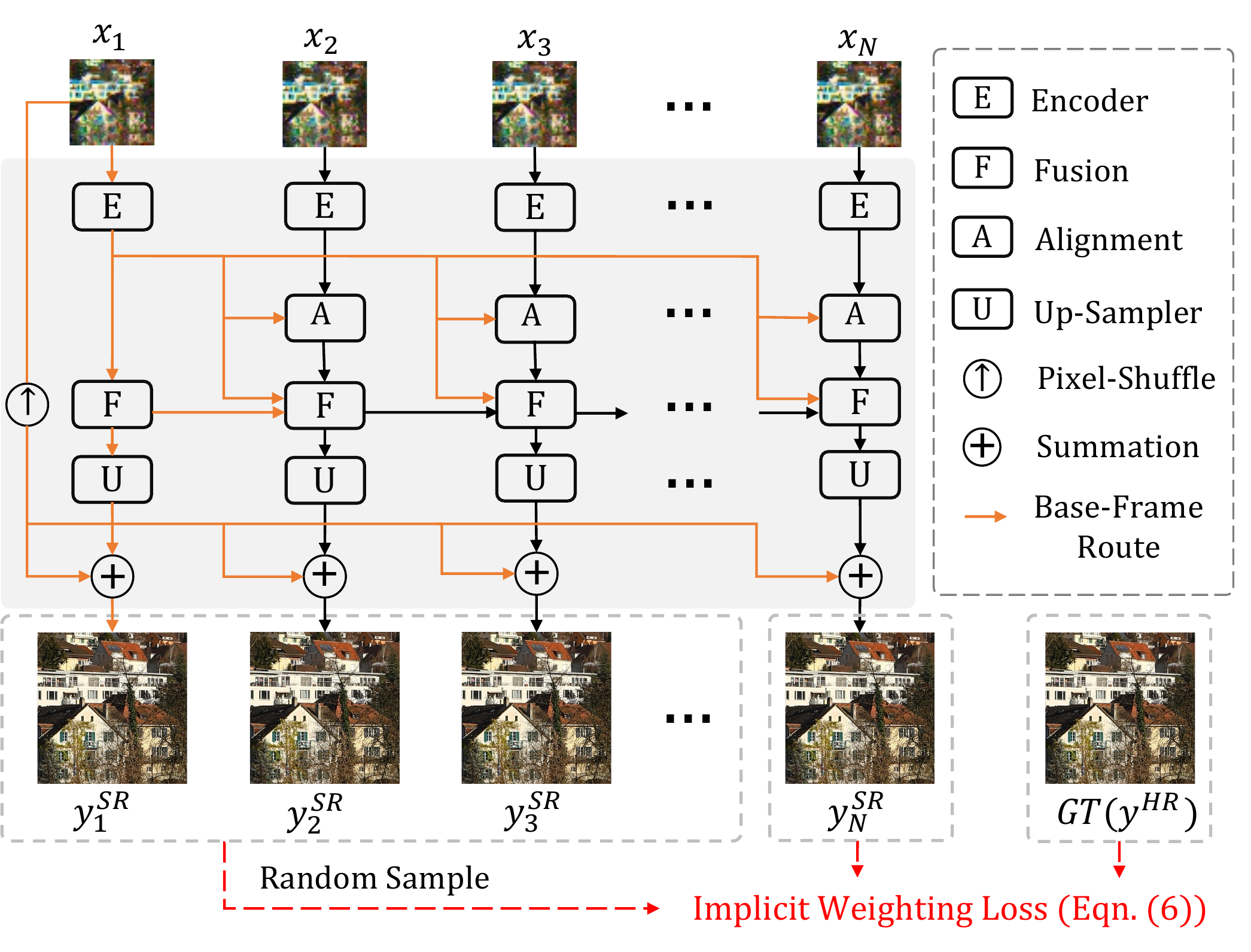}
    \vspace{-3mm}
    \caption{The architecture of the proposed RBSR.
    Each LR frame is first fed into the encoder to extract features which are then warped to the base ones by the alignment module.
    Next, the previous merged features and current aligned ones are passed into our proposed recurrent fusion module, which is shown in Fig.~\ref{fig:KFGR}. 
    Finally, the up-sampler is employed to generate the SR result. 
    Moreover, an implicit weighting loss is present to enhance the model's flexibility in processing input frames with variable lengths.}
    \label{fig:fugure2}
    \vspace{-6mm}
\end{figure*}
 
\noindent{\textbf{Recurrent Neural Network.}
Recurrent Neural Network(RNN)\cite{schuster1997bidirectional} provides an elegant way for temporal modeling and can be naturally applied to multi-frame restoration\cite{chen2016deep,huang2017video,fuoli2019efficient,chan2021basicvsr,chan2022basicvsr++,rong2020burst,wang2023benchmark,li2022efficient}. 
For instance, BasicVSR \cite{chan2021basicvsr} employs bidirectional propagation, achieving high performance. 
BasicVSR++\cite{chan2022basicvsr++} further enhances BasicVSR\cite{chan2021basicvsr} with second-order bidirectional propagation. 
Rong \etal.\cite{rong2020burst} proposes a temporally shifted wavelet transform to combine spatial and frequency information for burst denoising. 
In our work, we explore a specific recurrent manner suitable for BurstSR, and propose a compact, efficient, and flexible recurrent neural network RBSR.

\section{Method}

\subsection{Problem Formation and Pipeline}
\label{section:1}
\noindent \textbf{Problem Formation.} 
BurstSR aims at generating a clean and high-resolution sRGB image $y^{SR}_{N} \in \mathbb{R}^{sH \times sW \times 3}$ from multiple low-resolution and noisy RAW images ${\{x_i\}}^{N}_{i=1}$, where $x_i \in \mathbb{R}^{H \times W}$. 
$N$ and $s$ denote the input frame number and super-resolution factor, respectively.
Inter-frame alignment and feature fusion are two important aspects of BurstSR.
Existing methods have more fully explored the former~\cite{chan2021basicvsr,bhat2021deep,bhat2021mfir,wang2019edvr,luo2022bsrt,mehta2023gated,liang2022recurrent}, but are limited in the effectiveness~\cite{bhat2021deep,bhat2021mfir}, efficiency~\cite{luo2021ebsr,mehta2023gated,dudhane2023burstormer}, and flexibility~\cite{dudhane2022burst} of the latter.
In this work, we expect to explore a fusion method that is efficient and adaptive to variable frame numbers. 
Specifically, we suggest merging features frame-by-frame to utilize the beneficial information of each frame as much as possible, and propose an efficient and flexible recurrent network named RBSR.

\begin{figure}[t!]
    \centering
    \includegraphics[width=0.9\linewidth]{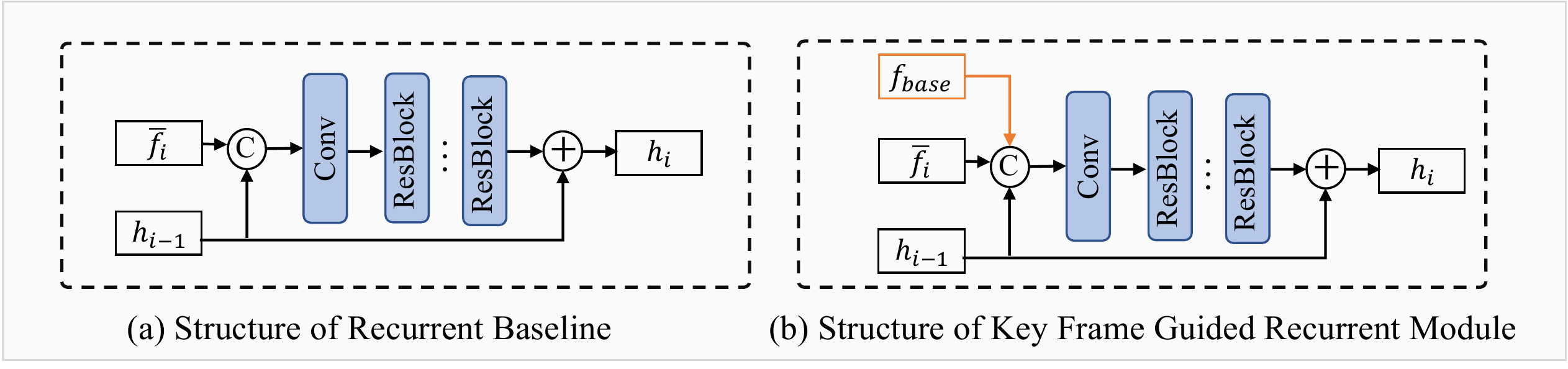}
    \caption{Structure of the proposed recurrent baseline and key frame guided recurrent (KFGR) module.}
    \label{fig:KFGR}
    \vspace{-6mm}
\end{figure}

\noindent \textbf{Pipeline.}
RBSR includes the encoder, alignment, recurrent fusion, and up-sampler module, as show in Fig. \ref{fig:fugure2}.
First, LR images are fed into the encoder $\mathcal{E}$ to obtain deep feature representations $\{f_i\}_{i=1}^{N}$.
To merge the features, the inevitable spatial misalignment issue between frames should be pre-addressed.
Thus, we deploy the alignment model $\mathcal{A}$ based on flow-guided deformable strategy~\cite{chan2022basicvsr++} to warp the non-base features to the base ones $f_{1}$, which can be written as,
\begin{equation}
\bar{f}_i = \mathcal{A}(f_{base}, f_i),
\end{equation}
where $f_{base}=f_{1}$, $\bar{f}_i$ denotes the aligned $f_i$.
Then, the recurrent fusion module $\mathcal{F}$ combines the aligned information frame-by-frame, which can be written as,
\begin{equation}\label{fusion}
{h}_N = \mathcal{F}(f_{base}, \bar{f}_2, ..., \bar{f}_N).
\end{equation}
Finally, the merged feature ${h}_N$ is taken into up-sampler to obtain HR result $y_{N}^{SR}$.

In this paper, we focus on the fusion manner and propose a key frame guided recurrent (KFGR) module in Sec. \ref{section:2}.
In comparison with common recurrent networks~\cite{chan2021basicvsr,chan2022basicvsr++}, KFGR utilizes the base-frame as a key prompt to guide the acquisition of knowledge from other frames.
Moreover, although recurrent networks can naturally generalize to variable input frame numbers, simply constraining the final output $y_{N}^{SR}$ limits the model flexibility, dropping the performance in handling shorter input sequences. 
To address this issue, in Sec. \ref{section:3}, we propose an implicit weighting loss, which enhances the ability in processing shorter sequences while maintaining performance on the longest one.

\subsection{Recurrent Fusion Module}
\label{section:2}
In this sub-section, we first design our basic recurrent fusion scheme. 
Then we further enhance the scheme by utilizing the base-frame more sufficiently, proposing a key frame guided recurrent (KFGR) module. 

\noindent \textbf{Recurrent Scheme.}
The base-frame mainly serves as spatial position guidance in BurstSR. 
However, the non-base features processed by the alignment model generally are not perfectly aligned with the base-frame, since the misalignment cases are varied and complex.
Such an inherent problem drives the fusion module to consider the base-frame as guidance, which can ensure that the merged features will not have position shifts, alleviating artifacts.
Thus, in this work, the recurrent fusion scheme is designed  to propagate information from the base-frame to others, rather than from others to the base-frame. The scheme can make the base-frame provide implicit guidance for others in every recurrence, and it can be also seen as a gradual restoration of the base-frame. 
Specifically, we consider a recurrent network incorporating residual learning strategy with two inputs: the current aligned feature $\bar{f}_i$ and the previous merged feature ${h}_{i-1}$, \ie,
\begin{equation}
\label{eqn:f—base}
{h}_i = \mathcal{F}({h}_{i-1}, \bar{f}_i) + {h}_{i-1}.
\end{equation}
where $\mathcal{F}$ is recurrent baseline module.
${h}_N$ will merge all frame information, and then be passed to the up-sampler for generating the final result.
For simplicity and efficiency, after concatenating inputs along channel dimension, we only use a $3\times3$ convolutional layer for channel reduction and several residual blocks~\cite{he2016deep} for feature enhancement, as illustrated in Fig. \ref{fig:KFGR} (a).

\noindent \textbf{Key Frame Guided Recurrent Module.}
In the above recurrent fusion scheme, the base-frame is only utilized directly at the beginning.
As the propagation progresses, the valuable information from base-frame may be lost, resulting in a weakened guiding effect.
Thus, we additionally provide the feature of base-frame as a key condition to guide the acquisition of knowledge from other frames in every recurrence.
Eqn.~(\ref{eqn:f—base}) can be modified as,
\begin{equation}
{h}_i = \mathcal{F}({h}_{i-1}, \bar{f}_i | f_{base}) + {h}_{i-1},
\end{equation}
where $\mathcal{F}$ is KFGR module. Without bells and whistles, KFGR simply concatenates the base-frame feature $f_{base}$, the current aligned feature $\bar{f}_i$, and the previous merged feature ${h}_{i-1}$ along channel dimension, as shown in Fig. \ref{fig:KFGR} (b).
And the backbone of KFGR remains the same as that of the recurrent baseline.
Such a more explicit scheme can enable more sufficient utilization of the base-frame, thereby improving the performance.

\noindent \textbf{Discussion.}
There are several advantages of our proposed recurrent fusion module KFGR.
First, merging cues frame-by-frame makes it effective, achieving favorable performance. 
Moreover, our experiments in Sec.~\ref{section:43} show that such a unidirectional propagation is more suitable than a bidirectional one for BurstSR.
Second, the compact network design makes it efficient, costing low inference time.
Third, the recurrent manner makes it flexible to deal with input frames of varying lengths, which is detailed in the next section.

\subsection{Implicit Weighting Loss}
\label{section:3}
Existing BurstSR methods\cite{bhat2021deep,bhat2021mfir,dudhane2022burst,mehta2023gated,dudhane2023burstormer,luo2021ebsr} are generally optimized by minimizing the  $\ell_1$ distance between the output $y_{N}^{SR}$ and the ground truth $y^{HR}$,  \ie,
\begin{equation}\label{L1loss}
\mathcal{L}=\left\|y_N^{S R} -y^{H R}\right\|_1.
\end{equation}
However, Eqn.~(\ref{L1loss}) only constrains the result with the input of all frames, and it leads to a significant performance decrease in facing fewer ones.
One straightforward solution is to apply loss in every recurrence, but it increases training time and weakens the ability to handle the longest frames.
The related experiments will be shown in Sec.~\ref{section:43}. 

To address the above issues, we propose an implicit weighting loss, which keeps the constraint in the last recurrence and randomly adds one in the previous recurrences. The loss can be written as,
\begin{equation}\label{weightingloss}
\begin{aligned}
\mathcal{L}=\left\|y_i^{S R}-y^{H R}\right\|_1 &+ \left\|y_N^{S R} -y^{H R}\right\|_1,
\end{aligned}
\end{equation}
where $i$ is randomly and uniformly sampled from $1$ to $N-1$.
On the one hand, the proposed loss implicitly imposes constraints in every recurrence, thus bringing performance improvement in facing fewer frames.
On the other hand, it also implicitly applies the strongest weight to the last result, thus making performance maintenance in processing the longest frames.
In addition, the implicit weighting loss only slightly increases the training cost compared with Eqn.~(\ref{L1loss}).

\section{Experiments}

\subsection{Experimental Settings} 
\label{section:41}

\noindent \textbf{Datasets.}
The experiments are conducted on synthetic SyntheticBurst and real-world BurstSR datasets\cite{bhat2021deep}.
SyntheticBurst dataset consists of 46839 bursts for training and 300 for testing. 
Each burst contains 14 LR RAW images that are synthesized from a single sRGB image captured by a Canon camera. 
Specifically, the sRGB image is first converted to the linear RGB values using an inverse camera pipeline\cite{brooks2019unprocessing}. 
Next, random translations and rotations are applied to generate a shifted burst. 
Then the transformed images are downsampled by the bilinear kernel and added noise, obtaining the low-resolution and noisy burst. 
Finally, Bayer mosaicking is deployed to get the input RAW burst.
BurstSR dataset involves 5405 patches for training, and 882 patches for testing.
The LR RAW images and the corresponding HR sRGB image are captured with a Samsung smartphone camera and a Canon DSLR camera, respectively. 

\noindent \textbf{Implement Details.}
The scale factor $s$ for super-resolution is set to $4$.
The RAW burst images are packed into four channels (\ie, RGGB) as inputs according to the Bayer filter pattern.
The batch size is set to 16. The input patch size is set to $48\times48$ for synthetic experiments and $56\times56$ for real-world experiments.
For the SyntheticBurst dataset\cite{bhat2021deep}, the model is trained with AdamW optimizer\cite{loshchilov2017decoupled} with $\beta_1=0.9$ and $\beta_2=0.999$ for 400k iterations.
Cosine annealing strategy\cite{loshchilov2016sgdr} is employed to steadily decrease the learning rate from $10^{-4}$ to $10^{-6}$.
For the BurstSR dataset\cite{bhat2021deep}, following \cite{bhat2021deep,bhat2021mfir,dudhane2022burst}, we fine-tune the pre-trained model on SyntheticBurst dataset for additional 50k iterations.
All experiments are conducted with PyTorch~\cite{paszke2019pytorch} on an Nvidia GeForce RTX 2080Ti GPU.

\begin{figure}[t!]
\centering
 \begin{overpic}[width=0.99\textwidth,grid=False]{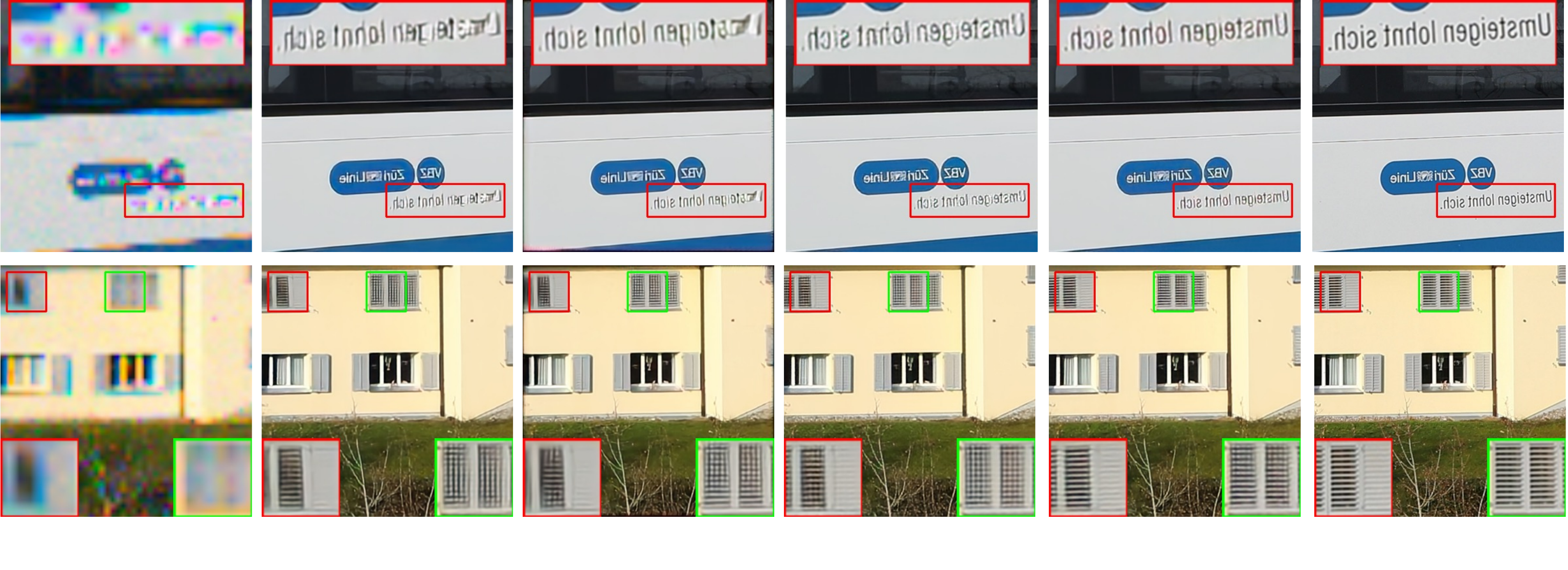}
  \put(25,1.5){\small{LR}}
  \put(65,1.5){\small{DBSR\cite{bhat2021deep}}}
  \put(125,1.5){\small{MFIR\cite{bhat2021mfir}}}
  \put(180,1.5){\small{BIPNet\cite{dudhane2022burst}}}
  \put(250,1.5){\small{Ours}}
  \put(310,1.5){\small{HR}}
\end{overpic}
\vspace{-2mm}
\caption{Qualitative comparison on synthetic SyntheticBurst dataset\cite{bhat2021deep}.}
\label{fig:syn_visualize}
\vspace{-3mm}
\end{figure}

\begin{figure}[t!]
\centering
 \begin{overpic}[width=0.99\textwidth,grid=False]{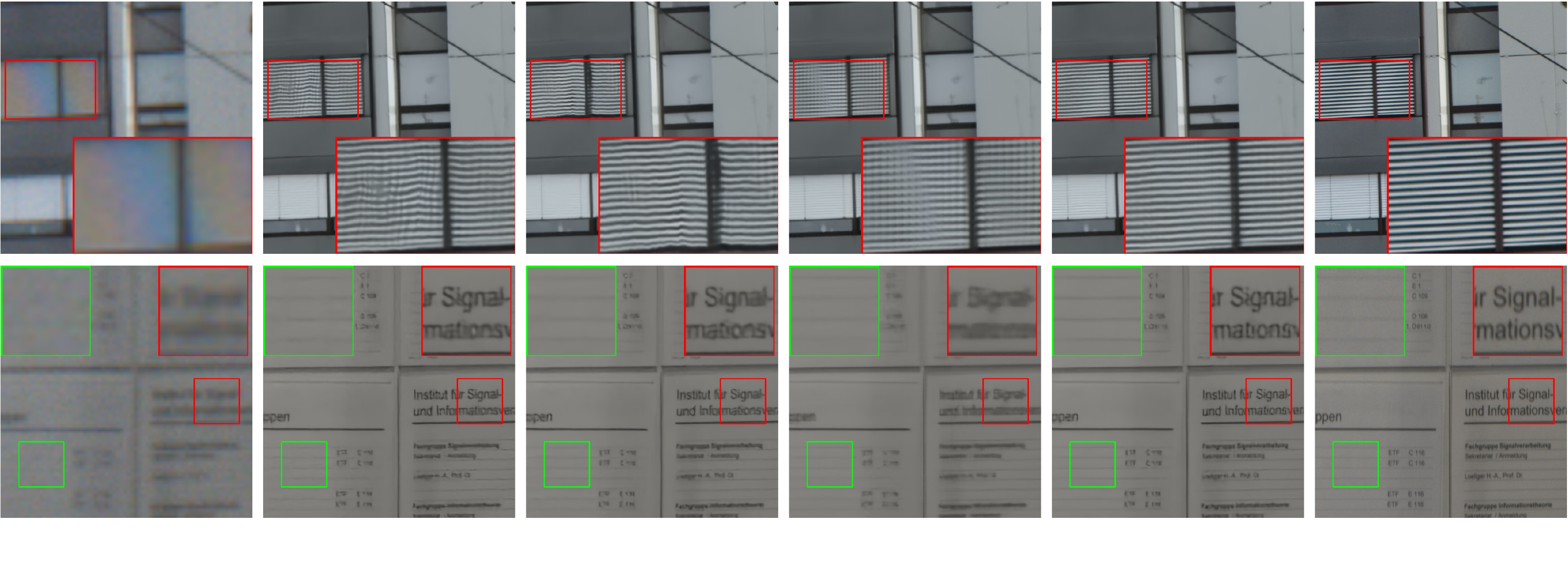}
  \put(25,1.5){\small{LR}}
  \put(65,1.5){\small{DBSR\cite{bhat2021deep}}}
  \put(125,1.5){\small{MFIR\cite{bhat2021mfir}}}
  \put(180,1.5){\small{BIPNet\cite{dudhane2022burst}}}
  \put(250,1.5){\small{Ours}}
  \put(310,1.5){\small{HR}}
\end{overpic}
\vspace{-2mm}
\caption{Qualitative comparison on real-world BurstSR  dataset\cite{bhat2021deep}.}
\label{fig:real_visualize}
\vspace{-5mm}
\end{figure}
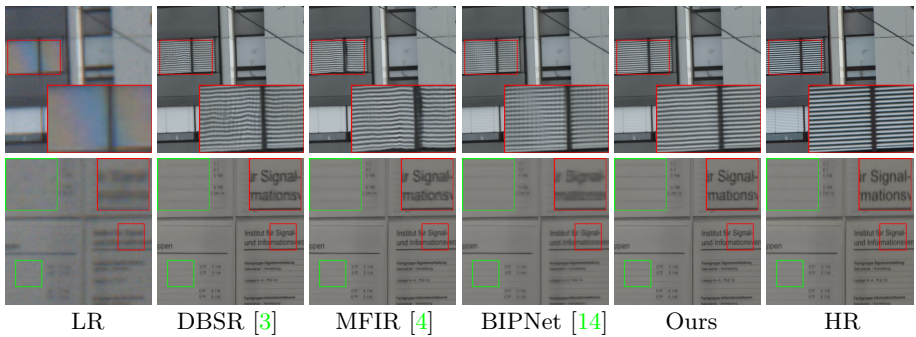

\begin{table}[t!]
\begin{center}
\small
\caption{Quantitative comparison on SyntheticBurst and BurstSR  datasets\cite{bhat2021deep}.}
\vspace{1mm}
\label{tab:comparison}
\setlength{\tabcolsep}{2mm}{
\begin{tabular}{cccccc}
\toprule
Method  & \tabincell{c}{SyntheticBurst \\ PSNR / SSIM } & \tabincell{c}{BurstSR \\ PSNR / SSIM }  & \tabincell{c}{\#Params \\ (M)} & \tabincell{c}{\#FLOPs \\ (G)} &\tabincell{c}{Time \\ (ms)}  \\
    \midrule
    HighResNet\cite{rong2020burst} & 37.45 / 0.924 &  46.64 / 0.980 & 34.78 &  96 & -\\
    DBSR\cite{bhat2021deep}     & 40.76 / 0.959 & 48.05 / 0.984 & 13.01 & 
    103 & {231}\\
    LKR\cite{lecouat2021lucas}     & 41.45 / 0.953 &  - & - & - & -\\
    MFIR\cite{bhat2021mfir}    & 41.56 / 0.964 &  48.33 / 0.985 & 12.13 & 121 & 428\\
    BIPNet\cite{dudhane2022burst}  & {41.93} / {0.967} & {48.49} / {0.985} & {6.66} & 326 & 691\\
    \midrule
    RBSR (Ours)   &{42.44} / {0.970} & {48.80} / {0.987} & {6.42} & 158 & {336}\\
\bottomrule
\end{tabular}
}
\end{center}
\vspace{-7mm}
\end{table}

\subsection{Comparison with State-of-the-Art Methods}
\label{section:42}

We compare the proposed RBSR with five BurstSR methods, \ie, HighResNet\cite{deudon2020highres}, DBSR\cite{bhat2021deep}, MFIR\cite{bhat2021mfir}, LKR\cite{lecouat2021lucas}, and BIPNet\cite{dudhane2022burst}.  
PSNR and SSIM\cite{hore2010image} are adopted as evaluation metrics. 
Additionally, the number of model parameters, \#FLOPs and inference time are also reported when generating a $1568 \times 1024$ HR image.

\noindent \textbf{Results on synthetic SyntheticBurst 
 dataset. } 
Table \ref{tab:comparison} shows the quantitative results.
It shows that our RBSR performs satisfactorily both in effectiveness and efficiency. 
In comparison with the recent state-of-the-art method BIPNet\cite{dudhane2022burst}, RBSR yields a PSNR gain of 0.51dB without increasing the number of parameters.
Besides, RBSR runs much faster, benefiting from concise and efficient design.
The qualitative results in Fig. \ref{fig:syn_visualize} show that our RBSR restores more rich and realistic textures and fewer artifacts than others.

\noindent \textbf{Results on real-world BurstSR dataset. }
The LRs and HR in the real-world dataset are slightly misaligned, since they are captured with different cameras in BurstSR dataset\cite{bhat2021deep}. 
To overcome this issue, following previous works\cite{bhat2021deep,bhat2021mfir,dudhane2022burst}, we train RBSR with the aligned $\ell_1$ loss and evaluate with the aligned PSNR as well as SSIM metrics in Table \ref{tab:comparison}.
In comparison with BIPNet, we achieve 0.64\% PSNR gains.
The qualitative results are shown in Fig. \ref{fig:real_visualize}. 
And the results demonstrate that our model still performs better, and recovers more details in the real world.

\noindent \textbf{Handling inputs with variable frame lengths.}
We compare with DBSR\cite{bhat2021deep} and MFIR\cite{bhat2021mfir} to verify our method's flexibility in facing inputs with variable frame numbers.
We did not compare with BIPNet\cite{dudhane2022burst} as it is only used in the case with a fixed input frame number. 
Table \ref{tab:comparisonframe} shows that our method significantly outperforms others across all numbers of input frames.
Especially, for our RBSR, taking 6 frames as inputs can obtain a similar performance with DBSR\cite{bhat2021deep} that is fed 14 frames.

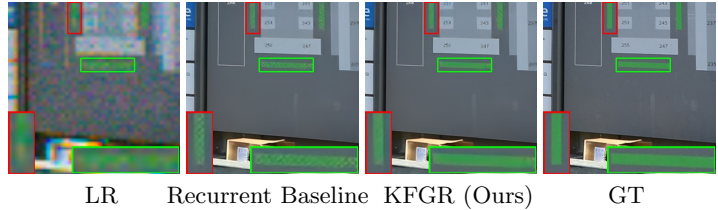
\begin{figure}[t!]
\centering
 \begin{overpic}[width=0.8\textwidth,grid=False]{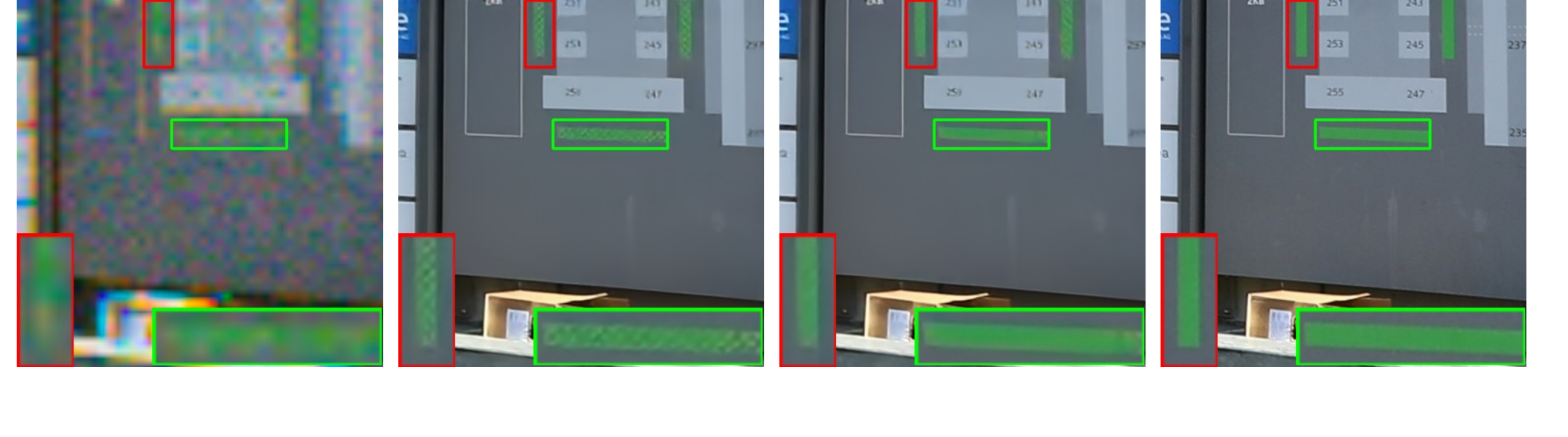}
  \put(32,1.5){\small{LR}}
  \put(63,1.5){\small{Recurrent Baseline}}
  \put(145,1.5){\small{KFGR (Ours)}}
  \put(230,1.5){\small{GT}}
\end{overpic}
\vspace{-2mm}
\caption{The result comparison when taking recurrent baseline and KFGR module (see Fig.~\ref{fig:KFGR}) as fusion manner.}
\label{fig:CatBase}
\vspace{-6mm}
\end{figure}

\begin{table}[t!]
\begin{center}
\caption{Qualitative comparison with other approaches when facing variable input frame numbers. We show `PSNR/SSIM' metrics in the table.}
\vspace{1mm}
\label{tab:comparisonframe}
\scriptsize
\setlength{\tabcolsep}{0.7mm}{
\begin{tabular}{cccccccc}
\toprule
Method  & N=2 & N=4  & N=6 & N=8 & N=10 & N=12  & N=14  \\
    \midrule
    DBSR\cite{bhat2021deep} & 34.78/0.892 & 37.42/0.930 & 38.90/0.945 & 39.73/0.952 & 40.22/0.956 & 40.55/0.958 & 40.78/0.959\\
    MFIR\cite{bhat2021mfir} & 36.37/0.915 & 38.70/0.942 & 39.81/0.952 & 40.50/0.958 & 40.98/0.961 & 41.32/0.963 & 41.55/0.964\\
    \midrule
    Ours & 38.08/0.935 & 39.85/0.952 & 40.79/0.959 & 41.41/0.963 & 41.86/0.966 & 42.19/0.968 & 42.44/0.970\\
\bottomrule
\end{tabular}
}
\end{center}
\vspace{-8mm}
\end{table}

\begin{table}[t]
\begin{center}
\small
\caption{Qualitative comparison of different loss  calculation approaches. We show ‘PSNR / SSIM’ metrics in the table.}
\vspace{1mm}
\label{tab:loss_comparison}
\setlength{\tabcolsep}{2mm}{
    \begin{tabular}{c c c c }
    \toprule
    Method & N=6 & N=10 & N=14\\
    \midrule
    Constraining Last Result & 39.79 / 0.953 & 41.53 / 0.965 & 42.45 / 0.970\\
    Constraining Every Result & 40.81 / 0.959 & 41.80 / 0.966 & 42.35 / 0.969\\
    \midrule
    Implicit Weighting Loss (Ours) & 40.79 / 0.959 & 41.86 / 0.966 & 42.44 / 0.970\\
    \bottomrule
\end{tabular}}
\vspace{-8mm}
\end{center}
\end{table}

\subsection{Ablation Study}
\label{section:43}
\noindent \textbf{Comparison of unidirectional and bidirectional propagation.}
We compare our unidirectional recurrent fusion model with two typical propagation schemes, \ie, bidirectional \cite{chan2021basicvsr} and enhanced bidirectional propagation \cite{chan2022basicvsr++}.
Although they have been widely used in video-restoration tasks, they have not shown any superiority for BurstSR, only receiving 42.34dB and 42.33dB PSNR respectively, while our RBSR achieves 42.44dB on SyntheticBurst dataset\cite{bhat2021deep}.

\noindent \textbf{Effect of base-frame utilization.}
Our sufficient utilization of the base-frame is reflected in two aspects: propagating information starting from the base-frame to others and providing the base-frame feature as an explicit prompt in KFGR module.
Here we conduct an experiment by propagating information in a reversed order, which leads 0.09dB PSNR drop.
Moreover, when we replace the recurrent baseline with KFGR module (see Fig.~\ref{fig:KFGR}), it enables 0.12dB PSNR gain.
Fig. \ref{fig:CatBase} also shows that the utilization of the base-frame can significantly reduce artifacts.

\noindent \textbf{Effect of implicit weighting loss.}
We compare our implicit weighting loss with two loss calculation approaches.
One only constrains the last result as shown in Eqn. (\ref{L1loss}), and the other constrains the results in every recurrence.
As shown in Table \ref{tab:loss_comparison}, our implicit weighting loss can improve performance for shorter sequences while maintaining performance for the longest ones.

\section{Conclusion}
BurstSR aims at reconstructing a high-resolution sRGB image from a sequence of low-resolution and noisy RAW images.
In this work, we focus on the multi-frame fusion manner, proposing an efficient and flexible recurrent model for BurstSR. 
On the one hand, the model merges cues frame-by-frame and emphatically utilizes the base-frame to guide the acquisition of knowledge from other frames in every recurrence.
On the other hand, implicit weighting loss is present to enhance the model's flexibility in processing inputs with variable lengths.
Experiments on both synthetic and real-world datasets demonstrate our method achieves better results than state-of-the-art ones.

\clearpage

\title{Supplementary Material}
\author{}
\authorrunning{Wu. et al.}
\institute{}
\maketitle

\renewcommand{\thesection}{\Alph{section}}
\renewcommand{\thetable}{\Alph{table}}
\renewcommand{\thefigure}{\Alph{figure}}
\renewcommand{\theequation}{\Alph{equation}}

\section{Contents }
The content of this supplementary material involves:
\begin{itemize}
\item Network Details in Sec.~\ref{section:1}
\item Evaluation Metrics in Sec.~\ref{section:2}
\item Effect of Residual Block Numbers in Sec.~\ref{section:3}
\item Effect of Alignment Strategy in Sec.~\ref{section:4}
\item Comparison with More Fusion Methods in Sec.~\ref{section:5}
\item More Qualitative Comparisons in Sec.~\ref{section:6}
\item Limitations in Sec.~\ref{section:7}
\end{itemize}

\section{Network Details }
\label{section:1}
The encoder of RBSR consists of a $3\times3$ convolutional layer and 5 residual blocks~\cite{he2016deep}.
The alignment module is deployed based on flow-guided deformable strategy~\cite{chan2022basicvsr++} to warp the non-base features to the base ones, as illustrated in Fig. \ref{fig:alignment}. 
The up-sampler consists of 5 residual blocks~\cite{he2016deep} and the pixel-shuffle operation with skip connections.
A final $3\times3$ convolutional layer is deployed in the tail of the up-sampler.

\section{Evaluation Metrics}
\label{section:2}
We adopt PSNR (Peak Signal-to-Noise Ratio) and SSIM (Structural Similarity) ~\cite{hore2010image} as evaluation metrics, following previous works~\cite{bhat2021deep,bhat2021mfir,dudhane2022burst}.
PSNR considers the pixel-wise error between a reference image $\mathbf{Y}$ and a reconstructed image $\mathbf{X}$, which can be expressed as,
\begin{equation}
\text{PSNR}=10 \cdot \mathbf{\log}_{10}\left(\frac{255^2}{\text{MSE}}\right). \\
\end{equation}
where $\cdot$ is the pixel-wise multiplication operation.
$\text{MSE}$ can be written as,
\begin{equation}
\text{MSE}=\frac{1}{H W} \sum_{i=0}^{H-1} \sum_{j=0}^{W-1}[\mathbf{X}(i, j)-\mathbf{Y}(i, j)]^2,
\end{equation}
where $H$ and $W$ denote the height and width of the image.

SSIM is based on the luminance difference $\text{L}(\mathbf{X},\mathbf{Y})$, contrast difference $\text{C}(\mathbf{X},\mathbf{Y})$, and structural difference $\text{S}(\mathbf{X},\mathbf{Y})$ between a reconstructed image $\mathbf{X}$ and a reference image $\mathbf{Y}$. It can be expressed as,
\begin{equation}
    \text{SSIM} = \text{L}(\mathbf{X},\mathbf{Y})^{\alpha} \cdot \text{C}(\mathbf{X},\mathbf{Y})^{\beta} \cdot \text{S}(\mathbf{X},\mathbf{Y})^{\gamma},
\end{equation}  
\begin{equation}
    \text{L}(\mathbf{X},\mathbf{Y}) = \frac{2 \mu_X \mu_Y + c_1}{\mu_X^2 + \mu_Y^2 + c_1},
\end{equation}
\begin{equation}
    \text{C}(\mathbf{X},\mathbf{Y}) = \frac{2 \sigma_X \sigma_Y + c_2}{\sigma_X^2 + \sigma_Y^2 + c_2},
\end{equation}
\begin{equation}
    \text{S}(\mathbf{X},\mathbf{Y}) = \frac{2 \sigma_{XY} + c_3}{\sigma_X \sigma_Y + c_3},
\end{equation}
where $\cdot$ is the pixel-wise multiplication operation.
$\mu_X$ and $\mu_Y$ are the mean value of $\mathbf{X}$ and $\mathbf{Y}$, respectively.
$\sigma_X$ and $\sigma_Y$ are the variance of $\mathbf{X}$ and $\mathbf{Y}$, respectively.
$\sigma_{XY}$ is the covariance between $\mathbf{X}$ and $\mathbf{Y}$. 
$c_1$, $c_2$, $c_3$, $\alpha$, $\beta$ and $\gamma$ are  constants.

\section{Effect of Residual Block Numbers}
\label{section:3}
We conduct ablation experiments with different number of residual blocks\cite{he2016deep} in key frame guided recurrent (KFGR) module.
The results can be seen in Table~\ref{tab:resblock_comparison}.
Naturally, more blocks generally bring more performance improvements.
To make a better trade-off between effectiveness and efficiency, we choose 40 residual blocks as default.

\begin{figure}[t!]
    \centering
    \includegraphics[width=0.6\linewidth]{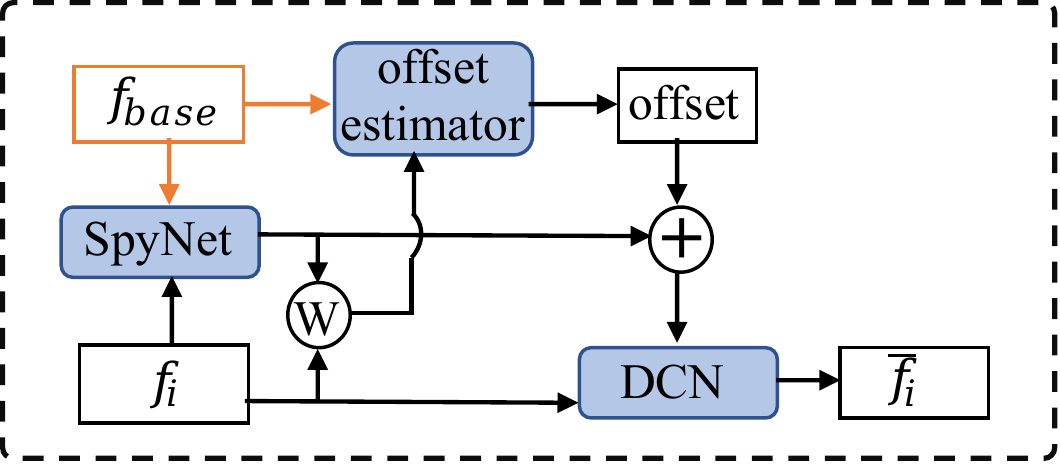}
    \caption{Structure of the alignment module based on flow-guided deformable strategy~\cite{chan2022basicvsr++}.}
    \vspace{-6mm}
    \label{fig:alignment}
\end{figure}

\begin{table}[t!]
\begin{center}
\caption{The results comparison on different residual block numbers.}
\label{tab:resblock_comparison}
\setlength{\tabcolsep}{3mm}{
\begin{tabular}{c c c c c}
\toprule
\multirow{2}{*}{$\#$Blocks}  & SyntheticBurst & BurstSR   &\multirow{2}{*}{Time (ms)} \\
 &  PSNR / SSIM &  PSNR / SSIM \\
\midrule
24   & 42.21 / 0.968 & 48.71 / 0.986  & 258  \\
32   & 42.36 / 0.969 & 48.76 / 0.987  & 293  \\
40   & 42.44 / 0.970 & 48.80 / 0.987  & 336  \\
48   & 42.50 / 0.970 & 48.79 / 0.987  & 375  \\
\bottomrule
\end{tabular}}
\end{center}
\vspace{-6mm}
\end{table}

\section{Effect of Alignment Strategy}
\label{section:4}
We conduct ablation experiments with different alignment strategies, including optical flow alignment, deformable alignment, and flow-guided deformable alignment~\cite{chan2022basicvsr++}.
SpyNet~\cite{ranjan2017optical} is adopted as the optical flow estimator.
The results on SyntheticBurst dataset~\cite{bhat2021deep} show that flow-guided deformable strategy enables 0.14 dB and 0.42 dB PSNR improvements compared to flow alignment and deformable alignment, respectively. 

\section{Comparison with More Fusion Methods}
\label{section:5}
We compare the proposed RBSR with five BurstSR methods, \ie, HighResNet\cite{deudon2020highres}, DBSR\cite{bhat2021deep}, MFIR\cite{bhat2021mfir}, LKR\cite{lecouat2021lucas}, and BIPNet\cite{dudhane2022burst}. 
Additionally, we further compare our proposed recurrent fusion method with the other two methods, the summation and the concatenation of all the input features, which get the fused feature maps by the summation and the concatenation of all input features followed with default 40 residual blocks, respectively.
Experiments on SyntheticBurst dataset~\cite{bhat2021deep} show that our proposed recurrent fusion method gets 0.93 dB and 0.81 dB PSNR gains compared with the summation and the concatenation of all the input features respectively. It further illustrates the effectiveness of our fusion method.

\section{More Qualitative Comparisons }
\label{section:6}
We perform more qualitative comparisons between the proposed RBSR and other BurstSR methods (\ie, DBSR\cite{bhat2021deep}, MFIR\cite{bhat2021mfir}, and BIPNet\cite{dudhane2022burst}).
The results on SyntehticBurst and BurstSR datasets\cite{bhat2021deep} are shown in Fig. \ref{fig:more_syn_visualize} and Fig. \ref{fig:more_real_visualize}, respectively.
For easy observation, we zoom out some areas within the red box.
It can be seen that our method can restore more fine-scale and photo-realistic textures than others, both on synthetic and real-world images.

\section{Limitations }
\label{section:7}
In low-light conditions, inevitable noise exists in the input images, and it influences the accuracy of the alignment module.
The inaccurate alignment makes it hard to effectively utilize the complementary information from input frames, leading to over-smoothing details in super-resolution results, as shown in Fig \ref{fig:failure_case}.

\clearpage

\begin{figure}[t!]
\centering
 \begin{overpic}[width=0.98\textwidth,grid=False]{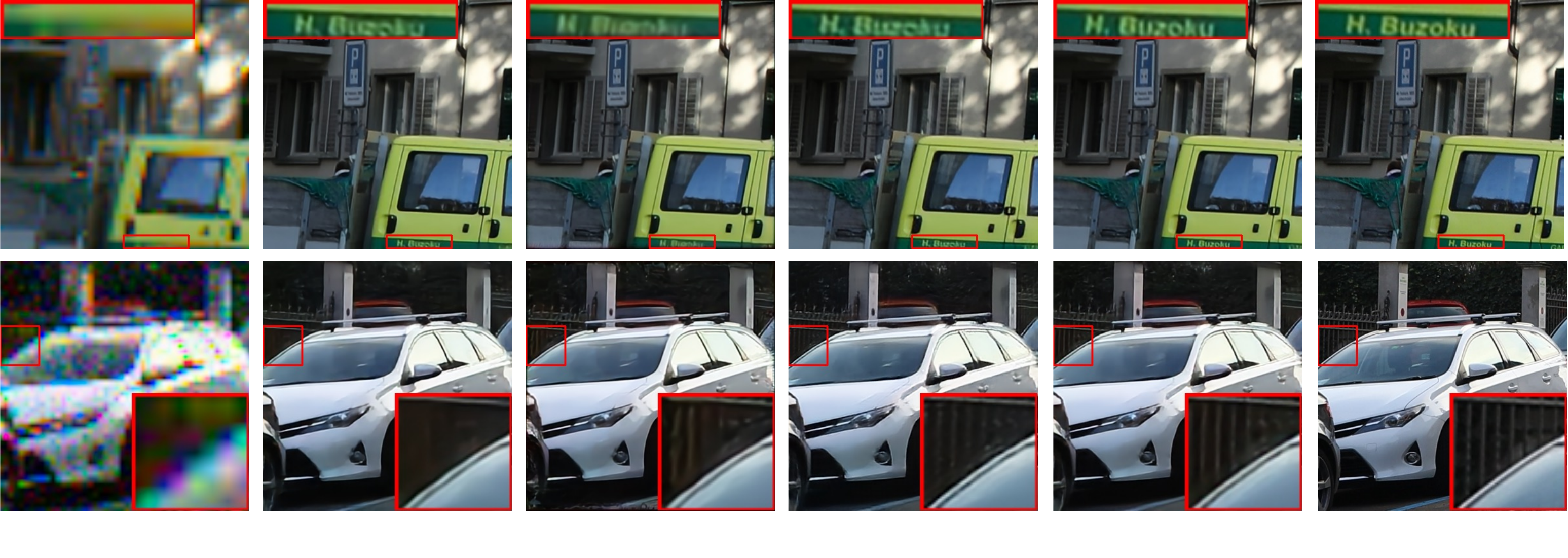}
  \put(20,1.5){\small{LR}}
  \put(65,1.5){\small{DBSR\cite{bhat2021deep}}}
  \put(120,1.5){\small{MFIR\cite{bhat2021mfir}}}
  \put(178,1.5){\small{BIPnet\cite{dudhane2022burst}}}
  \put(242,1.5){\small{Ours}}
  \put(298,1.5){\small{HR}}
\end{overpic}
\vspace{-1mm}
\caption{More qualitative comparison on synthetic SyntheticBurst dataset\cite{bhat2021deep}. Please zoom in for more details.}
\label{fig:more_syn_visualize}
\end{figure}

\begin{figure}[t!]
\centering
 \begin{overpic}[width=0.98\textwidth,grid=False]{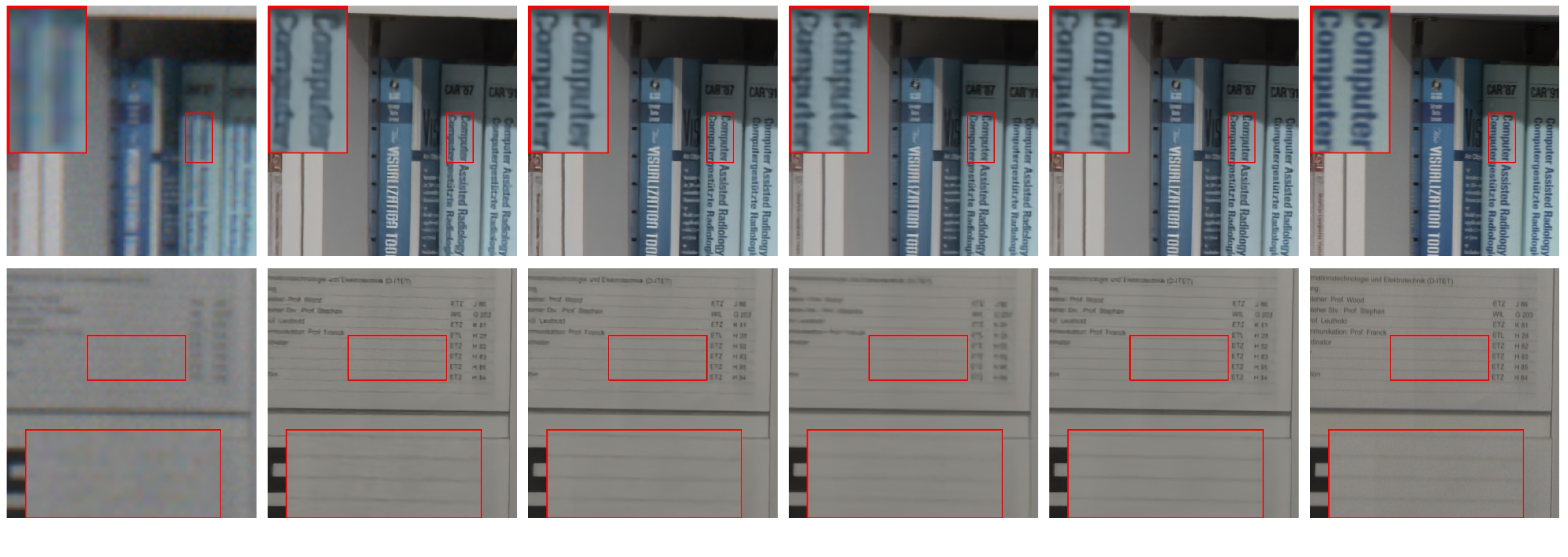}
  \put(20,0){\small{LR}}
  \put(65,0){\small{DBSR\cite{bhat2021deep}}}
  \put(120,0){\small{MFIR\cite{bhat2021mfir}}}
  \put(178,0){\small{BIPnet\cite{dudhane2022burst}}}
  \put(242,0){\small{Ours}}
  \put(298,0){\small{HR}}
\end{overpic}
\vspace{-1mm}
\caption{More qualitative comparison on real-world BurstSR dataset\cite{bhat2021deep}. Please zoom in for more details.}
\label{fig:more_real_visualize}
\end{figure}

\begin{figure}[t!]
\centering
 \begin{overpic}[width=0.98\textwidth,grid=False]{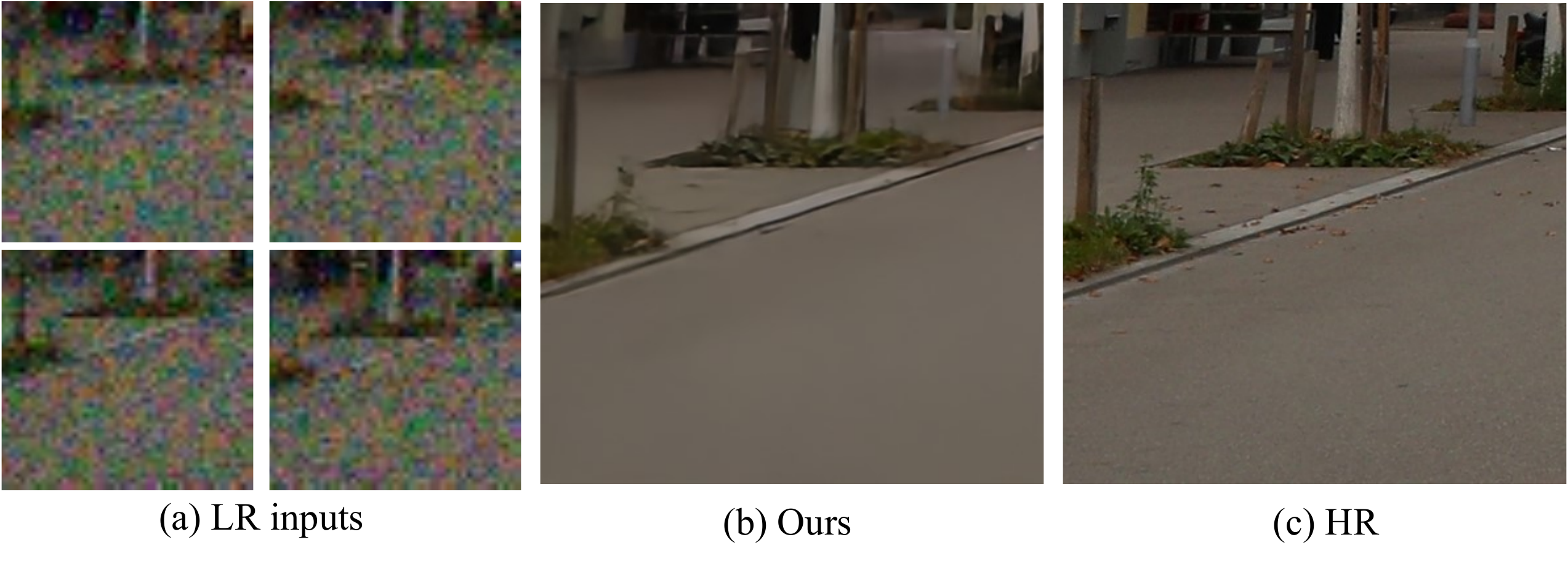}
\end{overpic}
\vspace{-1mm}
\caption{Results on synthetic SyntheticBurst dataset\cite{bhat2021deep}.
We use 14 LR frames as inputs but only visualize 4 ones.
Please zoom in for more details.}
\label{fig:failure_case}
\vspace{-1mm}
\end{figure}

\clearpage

\bibliographystyle{splncs04}
\bibliography{egbib}

\begin{thebibliography}{10}
\providecommand{\url}[1]{\texttt{#1}}
\providecommand{\urlprefix}{URL }
\providecommand{\doi}[1]{https://doi.org/#1}

\bibitem{bhat2021ntire}
Bhat, G., Danelljan, M., Timofte, R.: Ntire 2021 challenge on burst
  super-resolution: Methods and results. In: CVPRW (2021)

\bibitem{bhat2022ntire}
Bhat, G., Danelljan, M., Timofte, R., Cao, Y., Cao, Y., Chen, M., Chen, X.,
  Cheng, S., Dudhane, A., Fan, H., et~al.: Ntire 2022 burst super-resolution
  challenge. In: CVPRW (2022)

\bibitem{bhat2021deep}
Bhat, G., Danelljan, M., Van~Gool, L., Timofte, R.: Deep burst
  super-resolution. In: CVPR (2021)

\bibitem{bhat2021mfir}
Bhat, G., Danelljan, M., Yu, F., Van~Gool, L., Timofte, R.: Deep
  reparametrization of multi-frame super-resolution and denoising. In: CVPR
  (2021)

\bibitem{brooks2019unprocessing}
Brooks, T., Mildenhall, B., Xue, T., Chen, J., Sharlet, D., Barron, J.T.:
  Unprocessing images for learned raw denoising. In: CVPR (2019)

\bibitem{chan2021basicvsr}
Chan, K.C., Wang, X., Yu, K., Dong, C., Loy, C.C.: Basicvsr: The search for
  essential components in video super-resolution and beyond. In: CVPR (2021)

\bibitem{chan2022basicvsr++}
Chan, K.C., Zhou, S., Xu, X., Loy, C.C.: Basicvsr++: Improving video
  super-resolution with enhanced propagation and alignment. In: CVPR (2022)

\bibitem{chen2021pre}
Chen, H., Wang, Y., Guo, T., Xu, C., Deng, Y., Liu, Z., Ma, S., Xu, C., Xu, C.,
  Gao, W.: Pre-trained image processing transformer. In: CVPR (2021)

\bibitem{chen2016deep}
Chen, X., Song, L., Yang, X.: Deep rnns for video denoising. In: Applications
  of digital image processing. SPIE (2016)

\bibitem{dai2017deformable}
Dai, J., Qi, H., Xiong, Y., Li, Y., Zhang, G., Hu, H., Wei, Y.: Deformable
  convolutional networks. In: ICCV (2017)

\bibitem{delbracio2021mobile}
Delbracio, M., Kelly, D., Brown, M.S., Milanfar, P.: Mobile computational
  photography: A tour. Annual Review of Vision Science  (2021)

\bibitem{deudon2020highres}
Deudon, M., Kalaitzis, A., Goytom, I., Arefin, M.R., Lin, Z., Sankaran, K.,
  Michalski, V., Kahou, S.E., Cornebise, J., Bengio, Y.: Highres-net: Recursive
  fusion for multi-frame super-resolution of satellite imagery.
  arXiv:2002.06460  (2020)

\bibitem{dong2015image}
Dong, C., Loy, C.C., He, K., Tang, X.: Image super-resolution using deep
  convolutional networks. TPAMI  (2015)

\bibitem{dudhane2022burst}
Dudhane, A., Zamir, S.W., Khan, S., Khan, F.S., Yang, M.H.: Burst image
  restoration and enhancement. In: CVPR (2022)

\bibitem{dudhane2023burstormer}
Dudhane, A., Zamir, S.W., Khan, S., Khan, F.S., Yang, M.H.: Burstormer: Burst
  image restoration and enhancement transformer. CVPR  (2023)

\bibitem{fuoli2019efficient}
Fuoli, D., Gu, S., Timofte, R.: Efficient video super-resolution through
  recurrent latent space propagation. In: ICCVW. IEEE (2019)

\bibitem{he2016deep}
He, K., Zhang, X., Ren, S., Sun, J.: Deep residual learning for image
  recognition. In: CVPR (2016)

\bibitem{hore2010image}
Hore, A., Ziou, D.: Image quality metrics: Psnr vs. ssim. In: ICPR. IEEE (2010)

\bibitem{huang2017video}
Huang, Y., Wang, W., Wang, L.: Video super-resolution via bidirectional
  recurrent convolutional networks. TPAMI  (2017)

\bibitem{johnson2016perceptual}
Johnson, J., Alahi, A., Fei-Fei, L.: Perceptual losses for real-time style
  transfer and super-resolution. In: ECCV. Springer (2016)

\bibitem{kim2016accurate}
Kim, J., Lee, J.K., Lee, K.M.: Accurate image super-resolution using very deep
  convolutional networks. In: CVPR (2016)

\bibitem{kong2021classsr}
Kong, X., Zhao, H., Qiao, Y., Dong, C.: Classsr: A general framework to
  accelerate super-resolution networks by data characteristic. In: ECCV (2021)

\bibitem{lai2017deep}
Lai, W.S., Huang, J.B., Ahuja, N., Yang, M.H.: Deep laplacian pyramid networks
  for fast and accurate super-resolution. In: CVPR (2017)

\bibitem{lecouat2021lucas}
Lecouat, B., Ponce, J., Mairal, J.: Lucas-kanade reloaded: End-to-end
  super-resolution from raw image bursts. In: ICCV (2021)

\bibitem{li2022efficient}
Li, D., Zhang, Y., Law, K.L., Wang, X., Qin, H., Li, H.: Efficient burst raw
  denoising with variance stabilization and multi-frequency denoising network.
  IJCV  (2022)

\bibitem{li2023spatially}
Li, J., Zhang, Z., Liu, X., Feng, C., Wang, X., Lei, L., Zuo, W.: Spatially
  adaptive self-supervised learning for real-world image denoising. In: CVPR
  (2023)

\bibitem{liang2021swinir}
Liang, J., Cao, J., Sun, G., Zhang, K., Van~Gool, L., Timofte, R.: Swinir:
  Image restoration using swin transformer. In: ICCV (2021)

\bibitem{liang2022recurrent}
Liang, J., Fan, Y., Xiang, X., Ranjan, R., Ilg, E., Green, S., Cao, J., Zhang,
  K., Timofte, R., Gool, L.V.: Recurrent video restoration transformer with
  guided deformable attention. NeurIPS  (2022)

\bibitem{lim2017enhanced}
Lim, B., Son, S., Kim, H., Nah, S., Mu~Lee, K.: Enhanced deep residual networks
  for single image super-resolution. In: CVPRW (2017)

\bibitem{liu2020deep}
Liu, M., Zhang, Z., Hou, L., Zuo, W., Zhang, L.: Deep adaptive inference
  networks for single image super-resolution. In: ECCV. Springer (2020)

\bibitem{loshchilov2016sgdr}
Loshchilov, I., Hutter, F.: Sgdr: Stochastic gradient descent with warm
  restarts. arXiv:1608.03983  (2016)

\bibitem{loshchilov2017decoupled}
Loshchilov, I., Hutter, F.: Decoupled weight decay regularization.
  arXiv:1711.05101  (2017)

\bibitem{lugmayr2020srflow}
Lugmayr, A., Danelljan, M., Van~Gool, L., Timofte, R.: Srflow: Learning the
  super-resolution space with normalizing flow. In: ECCV. Springer (2020)

\bibitem{luo2022bsrt}
Luo, Z., Li, Y., Cheng, S., Yu, L., Wu, Q., Wen, Z., Fan, H., Sun, J., Liu, S.:
  Bsrt: Improving burst super-resolution with swin transformer and flow-guided
  deformable alignment. In: CVPR (2022)

\bibitem{luo2021ebsr}
Luo, Z., Yu, L., Mo, X., Li, Y., Jia, L., Fan, H., Sun, J., Liu, S.: Ebsr:
  Feature enhanced burst super-resolution with deformable alignment. In: CVPR
  (2021)

\bibitem{mehta2023gated}
Mehta, N., Dudhane, A., Murala, S., Zamir, S.W., Khan: Gated multi-resolution
  transfer network for burst restoration and enhancement. CVPR  (2023)

\bibitem{paszke2019pytorch}
Paszke, A., Gross, S., Massa, F., Lerer, A., Bradbury, J., et~al.: Pytorch: An
  imperative style, high-performance deep learning library. NeurIPS  (2019)

\bibitem{rong2020burst}
Rong, X., Demandolx, D., Matzen, K., Chatterjee, P., Tian, Y.: Burst denoising
  via temporally shifted wavelet transforms. In: ECCV. Springer (2020)

\bibitem{ronneberger2015u}
Ronneberger, O., Fischer, P., Brox, T.: U-net: Convolutional networks for
  biomedical image segmentation. In: MICCAI. Springer (2015)

\bibitem{schuster1997bidirectional}
Schuster, M., Paliwal, K.K.: Bidirectional recurrent neural networks. IEEE TSP
  (1997)

\bibitem{sun2018pwc}
Sun, D., Yang, X., Liu, M.Y., Kautz, J.: Pwc-net: Cnns for optical flow using
  pyramid, warping, and cost volume. In: CVPR (2018)

\bibitem{Tsai1984MultiframeIR}
Tsai, R.Y., Huang, T.S.: Multiframe image restoration and registration (1984)

\bibitem{vaswani2017attention}
Vaswani, A., Shazeer, N., Parmar, N., Uszkoreit, J., Jones, L., Gomez, A.N.,
  Kaiser, {\L}., Polosukhin, I.: Attention is all you need. NeurIPS  (2017)

\bibitem{wang2021exploring}
Wang, L., Dong, X., Wang, Y., Ying, X., Lin, Z., An, W., Guo, Y.: Exploring
  sparsity in image super-resolution for efficient inference. In: CVPR (2021)

\bibitem{wang2023benchmark}
Wang, R., Liu, X., Zhang, Z., Wu, X., Feng, C.M., Zhang, L., Zuo, W.: Benchmark
  dataset and effective inter-frame alignment for real-world video
  super-resolution. In: CVPRW (2023)

\bibitem{wang2019edvr}
Wang, X., Chan, K.C., Yu, K., Dong, C., Change~Loy, C.: Edvr: Video restoration
  with enhanced deformable convolutional networks. In: CVPRW (2019)

\bibitem{wang2021real}
Wang, X., Xie, L., Dong, C., Shan, Y.: Real-esrgan: Training real-world blind
  super-resolution with pure synthetic data. In: ICCV (2021)

\bibitem{wang2022uformer}
Wang, Z., Cun, X., Bao, J., Zhou, W., Liu, J., Li, H.: Uformer: A general
  u-shaped transformer for image restoration. In: CVPR (2022)

\bibitem{wei2020component}
Wei, P., Xie, Z., Lu, H., Zhan, Z., Ye, Q., Zuo, W., Lin, L.: Component
  divide-and-conquer for real-world image super-resolution. In: ECCV. Springer
  (2020)

\bibitem{wronski2019handheld}
Wronski, B., Garcia-Dorado, I., Ernst, M., Kelly, D., Krainin, M., Liang, C.K.,
  Levoy, M., Milanfar, P.: Handheld multi-frame super-resolution. TOG  (2019)

\bibitem{xie2021learning}
Xie, W., Song, D., Xu, C., Xu, C., Zhang, H., Wang, Y.: Learning
  frequency-aware dynamic network for efficient super-resolution. In: ICCV
  (2021)

\bibitem{zamir2022restormer}
Zamir, S.W., Arora, A., Khan, S., Hayat, M., Khan, F.S., Yang, M.H.: Restormer:
  Efficient transformer for high-resolution image restoration. In: CVPR (2022)

\bibitem{zhang2017beyond}
Zhang, K., Zuo, W., Chen, Y., Meng, D., Zhang, L.: Beyond a gaussian denoiser:
  Residual learning of deep cnn for image denoising. TIP  (2017)

\bibitem{zhang2018learning}
Zhang, K., Zuo, W., Zhang, L.: Learning a single convolutional super-resolution
  network for multiple degradations. In: CVPR (2018)

\bibitem{zhang2020texture}
Zhang, Y., Zhang, Z., DiVerdi, S., Wang, Z., Echevarria, J., Fu, Y.: Texture
  hallucination for large-factor painting super-resolution. In: ECCV. Springer
  (2020)

\bibitem{zhang2021learning}
Zhang, Z., Wang, H., Liu, M., Wang, R., Zhang, J., Zuo, W.: Learning
  raw-to-srgb mappings with inaccurately aligned supervision. In: ICCV (2021)

\end{thebibliography}
\end{document}